\documentclass{article}
\usepackage{graphicx} 
\usepackage[preprint]{dlde_neurips_2023}
\usepackage{amsmath}
\usepackage{amsfonts}
\usepackage{subfig}
\usepackage{natbib}
\usepackage{hyperref}
\usepackage{bm}

\title{One-Shot Transfer Learning for Nonlinear ODEs}

\author{
  Wanzhou Lei \\
  Harvard University \\
  \texttt{wanzhoulei@g.harvard.edu} \\
  \And
  Pavlos Protopapas \\
  Harvard University \\
  \texttt{pprotopapas@g.harvard.edu} 
  \And 
  Joy Parikh \\
  Columbia University \\
  jvp2118@columbia.edu
}

\begin{document}

\maketitle

\begin{abstract}

We introduce a generalizable approach that combines perturbation method and one-shot transfer learning to solve nonlinear ODEs with a single polynomial term, using Physics-Informed Neural Networks (PINNs). Our method transforms non-linear ODEs into linear ODE systems, trains a PINN across varied conditions, and offers a closed-form solution for new instances within the same non-linear ODE class. We demonstrate the effectiveness of this approach on the Duffing equation and suggest its applicability to similarly structured PDEs and ODE systems.

\end{abstract}

\section{Introduction}

Differential equations are crucial in scientific modeling, traditionally solved by methods such as Runge-Kutta and finite element analysis. Recently, Physics-Informed Neural Networks (PINNs) have shown promise in solving ODEs and PDEs by leveraging neural network capabilities (see \cite{hao2023physicsinformed}, \cite{Karniadakis2021}, \cite{lagaris_original_PINN}, \cite{NASCIMENTO2020103996}and reference within). However, computational cost remains a barrier, as PINNs cannot be generalized across different instances of similar equation types \cite{NEURIPS2021_df438e52}, \cite{fesser2023understanding}, a workaround is to train for multiple instances of same equation \cite{flamant2020solving}. To address this, we propose a novel hybrid approach that combines the perturbation method with one-shot transfer learning on PINNs \cite{protopapas2021} to efficiently and accurately solve non-linear ODEs of same type. 

\textbf{Related Work}
Introduced in 1998, neural networks for solving differential equations paved the way for today's Physics-Informed Neural Networks (PINNs) 
\cite{lagaris_original_PINN}, NeuroDiffEq and DeepXDE \cite{lu2019, chen2020} are two popular software programs that employ PINNs. PINNs have gained popularity and find applications in many domains today, a review of current state-of-the-art applications of PINNs can be found in \cite{hao2023physicsinformed}, \cite{bdcc6040140} and \cite{Karniadakis2021}.  PINNs have shown tremendous success in solving complex problems whose analytical solutions don't exist \cite{PhysRevD.107.063523}, however they perform poorly with generalizing solutions. Work related to adding physical constraints in NN structure \cite{mattheakis2020physical}, bounding errors on PINNs \cite{liu2023residualbased}, characterizing and mitigating failure modes \cite{NEURIPS2021_df438e52} and improving uncertainty quantification on Bayesian PINNs \cite{graf2022erroraware} has been important in increasing reliability of PINNs. PINNs handle interpolation problems very well, however face trouble with extrapolation. Transfer learning (TL) methods may be a potential solution, a study of effectiveness of TL can be found in this work \cite{fesser2023understanding} and an application can be found here \cite{pellegrin2022transfer}. In \citep{pellegrin2022transfer, zou2023lhydra}, a multi-headed neural network was used to learn the ``latent space" of a class of differential equations. Researchers have developed a transfer learning approach to solve linear differential equations in ``one-shot" \cite{protopapas2021}. Extending this to non-linear equations is not possible without modifications since non-linear equations have a non-quadratic loss function which cannot be optimized analytically in one-shot. The head weights have to be learned through an iterative process, such as gradient descent. Our study fills this gap in the literature by extending one-shot transfer learning to non-linear ODEs using perturbation method. 

\section{Methodology}
We are interested in solving non-linear ODEs with a single polynomial non-linear term of the following form:
\begin{equation}\label{ODEForm}
    Dx + \epsilon x^q = f(t).
\end{equation}

where $D$ is a differential operator of the form $D = \sum_{j=0}^m g_j \frac{d^j}{dt^j}$ and the RHS is a time-dependent forcing function. Note that we define $g_0 \frac{d^0}{dt^0} x = g_0 x$. The equation is also subject to a boundary conditions: $x(t=0) = x^{*}$ and $\frac{d^j}{dt^j} x(t=0) = x^{(j)*}$ for $j=1, 2, ..., m-1$. 
Our framework employs perturbation and one-shot transfer learning to solve a specific class of non-linear ODEs. We plan to extend this to handle systems of ODEs and PDEs.

\textbf{Perturbation Method} 

As mentioned above, non-linear ordinary differential equations (ODEs) do not possess analytical solutions to their loss functions with respect to the weights of the linear output layer, which is necessary for one-shot transfer learning.

In order to remove the non-linearity in the equation, we approximate the non-linear term $\epsilon x^q$ as a composition of functions \cite{PerturbMethod}. Assume $x = \sum_{i=0}^{\infty} \epsilon^i x_i$, where $x_i$ are unknown functions of t. We approximate $x$ with only p terms as: $x \approx \sum_{i=0}^{p} \epsilon^i x_i$. It is important to note that this truncated expansion of $x$ is only meaningful when the magnitude of $\epsilon$ is less than 1. Furthermore, the p-term approximation is more precise when the magnitude of $\epsilon$ is smaller. Fortunately, in most cases, we can adjust the equation by scaling to reduce the magnitude of $\epsilon$. When we substitute the p-term approximation into \ref{ODEForm} and expand using multinomial theorem , we obtain:



\begin{equation}
\sum_{i=0}^p \epsilon^i Dx_i + \epsilon \left( \sum_{k_0 + k_1 + ... + k_p = q}\frac{q!}{k_1!k_2!...k_p!}\epsilon^{\sum_{i=0}^p ik_i} \prod_{i=0}^p x_i^{k_i} \right) = f,
\end{equation}

The LHS of equation (3) is a polynomial of $\epsilon$. Since it holds for all values of $\epsilon$, the $0^{th}$ order term of LHS should be equal to the RHS and the coefficients of all higher-order terms of $\epsilon$ should all be 0. Therefore, for the $0^{th}$ order, we obtain: $Dx_0 = f$, and more generally for the $j^{th}$ order, where $1 \leq j \leq p$, we have:

\begin{equation}
    Dx_j = - \sum_{\substack{k_0 + k_1 +...+ k_p = q \\ \sum_{i=0}^p ik_i = j-1}}\frac{q!}{k_1!k_2!...k_p!} \prod_{i=0}^p x_i^{k_i} := f_j.
\end{equation}
where $f_j$ is the forcing function for $j^{th}$ ODE. The first few terms for this expansion look like:
\begin{equation}
\begin{aligned}
    Dx_0 &= f &
    Dx_1 &= -x_0^2 & 
    Dx_2 &= -2x_0x_1 &
    ...
\end{aligned}
\end{equation}
The forcing function $f_j$ depends only on previously solved $x_i$'s.  Therefore, (1) is reduced to a series of $p+1$ linear ODEs of the same form: $Dx_j = f_j$ that can be solved iteratively. There are a variety of ways to ensure that the initial boundary conditions are met. The main concept is to fix all $p+1$ boundary conditions to be the same, so that the total solution's, $x$, boundary condition is satisfied; that is, for all $k=0, 1, ...p$, $x_k(t=0) = x^*/\sum_{i=0}^p \epsilon^i$ and $\frac{d^j}{dt^j}x_k(t=0) = x^{(j)*}/\sum_{i=0}^p \epsilon^i$.

\textbf{Multi-head Fully Connected Neural Network} \\
As established earlier, solving the non-linear ODE is equivalent to solving a sequence of linear ODEs in the form \(Dx = f\). To minimize computational complexity, we transform higher-order differential equations into first-order equations by introducing $m-1$ additional dependent variables for an \(m^{th}\) order differential equation. Let \(\bm{u} = [x, x^{(1)}, x^{(2)}, \ldots, x^{(m-1)}]^T\) be a function mapping from \(\mathbb{R}\) to \(\mathbb{R}^m\), where \(x^{(1)} = \dot{x}\) and \(x^{(i)} = \dot{x}^{(i-1)}\) for \(i = 2, \ldots, m-1\). The equation \(Dx = f\) is then reduced to a first-order linear ODE system (see Appendix C for details).

\begin{equation}\label{sysode}
    \begin{cases}
        \dot{x} - x^{(1)} = 0 \\
        \dot{x}^{(i-1)} - x^{(i)} = 0, i=2, 3, 4, ..., m-1 \\
        g_0 x + \sum_{i=1}^{m-1}g_ix^{(i)} + g_m \dot{x}^{(m-1)} = f.
    \end{cases}
\end{equation}

Equation \ref{sysode} is equivalent to: $B\dot{\bm{u}} + A\bm{u} = F_j$ with boundary conditions $\bm{u}(t=0) = \bm{u}^* \in \mathbb{R}^m$, where $\dot{\bm{u}} = [\dot{x}, \dot{x}^{(1)}, \dot{x}^{(2)}, ..., \dot{x}^{(m-1)}]^T$ and $F_j = [0, 0, ..., f_j]^T$. A detailed description of the matrices $A$ and $B$ can be found in Appendix A. 

We create a fully connected neural network with $K$ heads in two parts to approximate the $K$ functions $\{\bm{u}_k \}_{k=1}^K$. The first part connects a 1D input to hidden layers, with the last layer having dimension $m h$. The activations of the last hidden layer are reshaped into a matrix $H \in \mathbb{R}^{m \times h}$, which reflects the hidden state of the ODE class and is then passed to the second part of the network. $H$ connects to $K$ heads, each associated with a linear ODE system. The output of each head is $\hat{\bm{u}}_k = HW_k \in \mathbb{R}^m$. A diagram of the general structure of the network can be found in Appendix B. The loss function for the $k^{th}$ head in the network is defined over a sampled data set $T$ as :

\begin{equation}\label{losseq}
    L_k = \frac{1}{mN}\sum_{t \in T} || B_k\dot{\hat{\bm{u}}}_k (t) + A_k\hat{\bm{u}}_k (t) - F_k(t) ||_2^2 + \frac{1}{m}||\hat{\bm{u}}_k (0) - \bm{u}_k^*||_2^2.
\end{equation}

where $\bm{u}_k^*$ is the boundary condition of the $k^{th}$ ODE. The total loss of the network is defined as: $L_{total} = \frac{1}{K}\sum_{k=1}^K L_k$ The purpose of training this neural network is to learn the latent space for one class of linear ODEs. Ideally, the larger K is, the better the learning of latent space and hence a generalization to a wider range of parameters.

\textbf{One-Shot Transfer Learning} \\
After training, we freeze the weights in the hidden layers. When encountering a new ODE of the same class, we only use one head, and the weights in this head can be calculated analytically in one shot.  Suppose $W$ is the time-independent network parameter in the last layer. The network now becomes: $\hat{\bm{u}}(t) = H(t)W$. We get the loss for this single-head neural network by substituting $\hat{\bm{u}}_k = \hat{\bm{u}}(t) = H(t)W$ into Eq.~\ref{losseq}:

\begin{equation}
\begin{split}
    L &= \frac{1}{mN}\sum_{t\in T}^n||B\dot{H}_t W + AH_tW - F(t)||_2^2 + \frac{1}{m}||H_0W - \bm{u}^*||_2^2,
\end{split}
\end{equation}

where $H_t$ is the hidden state of the network evaluated at $t$ and $H_0$ is the hidden state at the boundary. Differentiating $L$ from $W$ and setting $\frac{dL}{dW} = 0$, we obtain (details omitted):

\begin{eqnarray}\label{W}
    W &=& M^{-1}\left( H_0^T \bm{u}^* + \frac{1}{N}\sum_{t\in T} B\dot{H}_t F(t) + \frac{1}{N}\sum_{t\in T} H^T_t A^T F(t) \right), \\ 
      M &=& \frac{1}{N}\sum_{t\in T} (\dot{H}_t^T B^TB \dot{H}_t + \dot{H}_t^T B^TAH_t + H^T_tA^T B\dot{H}_t
    + H^T_t A^T A H_t) + H_0^T H_0.
\end{eqnarray}



For a fixed Duffing equation, the matrices $A$ and $B$ are fixed for all its $p+1$ reduced ODE systems. Thus, $M$ only needs to be computed and inverted once. We only need to update the forcing function $F$ in \ref{W} which iteratively depends on previous solutions. By reusing the first part of the neural network and using only one head, we optimally and iteratively compute the head parameters for each ODE system to solve them.

\section{Result}
We applied our proposed methodology to the 1D Duffing equation. The Duffing equation we are interested in \ref{duffingEq} is a second order non-linear ODE with five parameters: $\delta, \alpha, \beta, \gamma, \omega$ and one boundary condition $x(0) = x^*$. All higher-order boudnary conditions are set to 0. 

\begin{equation}\label{duffingEq}
    \frac{d^2 x}{dt^2} + \delta \frac{dx}{dt} + \alpha x + \beta x^3 = \gamma cos(\omega t).
\end{equation}

Using our framework, we first utilized the perturbation method and introduced new variables to reduce the Duffing equation to a series of $p+1$ first-order linear ODE systems of the form: $\dot{u}_i + Au_i = F_i$. We then built a network described in Section 2.2 with 10 heads. Each head represents a unique parameter setting. The specific details of the network structure can be found in Appendix B. The 10 parameter sets are uniformly randomly generated in the following range: 
 
 \begin{equation}\label{paraRange}
\gamma \in (0.5, 3), \omega \in (0.5, 3), \alpha \in (0.5, 4.5), \delta \in (0.5, 4.5), \bm{u}_1^* \in (-3, 3).
 \end{equation}

and $\bm{u}_2^* = 0$. After training (details in Appendix B), the network can accurately solve the 10 systems, and it acquires a significant understanding of the latent space of the not-linear ODE. We then test our method on an unseen Duffing equation. We measure the performance of the TL solution by computing the ODE loss of the Duffing equation. We used 14 different values of $p$ to solve and approximate the solution. As shown in Figure 1(a), as the $p$ value increases, the Duffing ODE loss decreases to around $10^{-3.75}$. The elbow shape can be used to figure out how many terms should be included in the perturbation expansion $p$.

\begin{figure}[htp] 
    \centering
    \subfloat[Log Duffing Equation Loss vs. p Values]{%
        \includegraphics[width=0.5\textwidth]{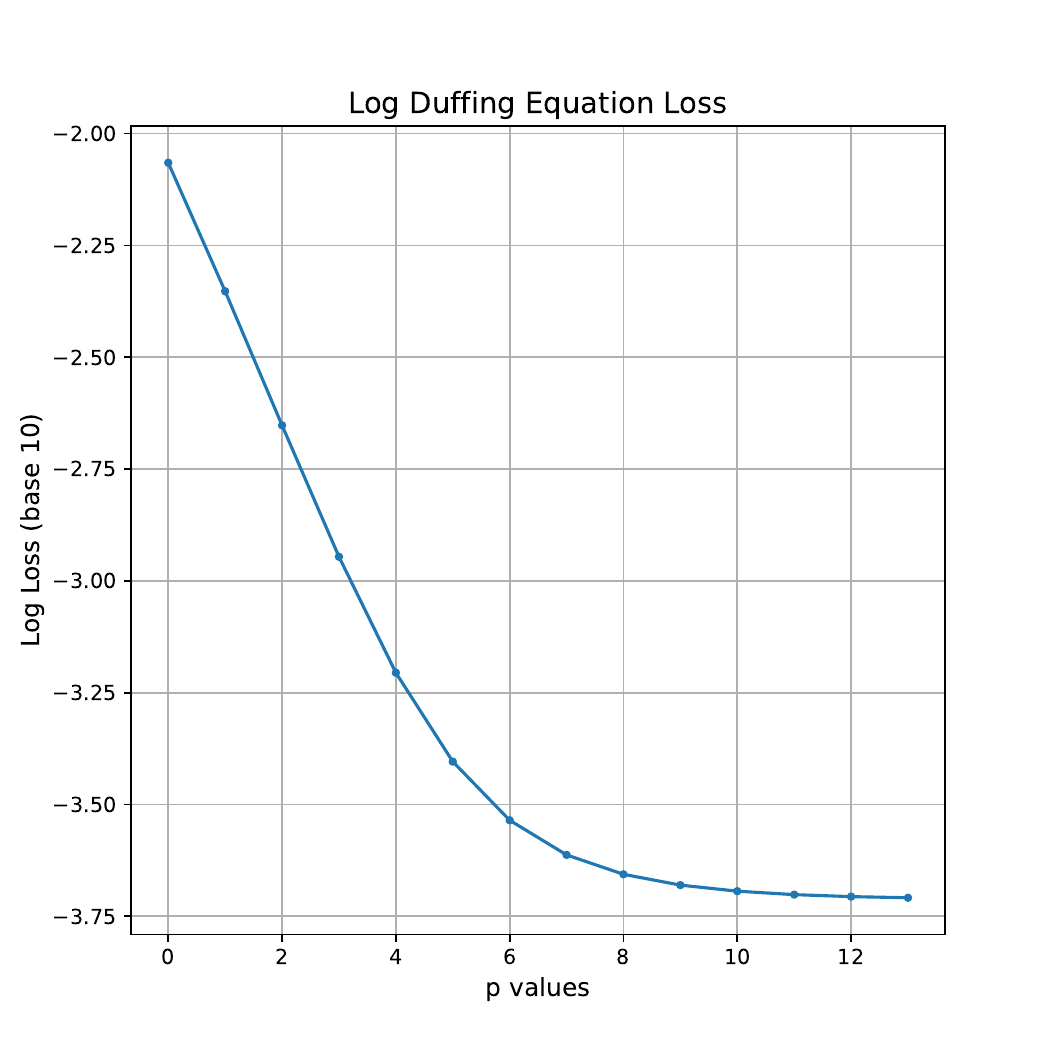}%
        \label{fig:a}%
        }
    \hfill%
    \subfloat[Numerical and TL Solutions of 20 Randon In-distribution Duffing Equation]{%
        \includegraphics[width=0.5\textwidth]{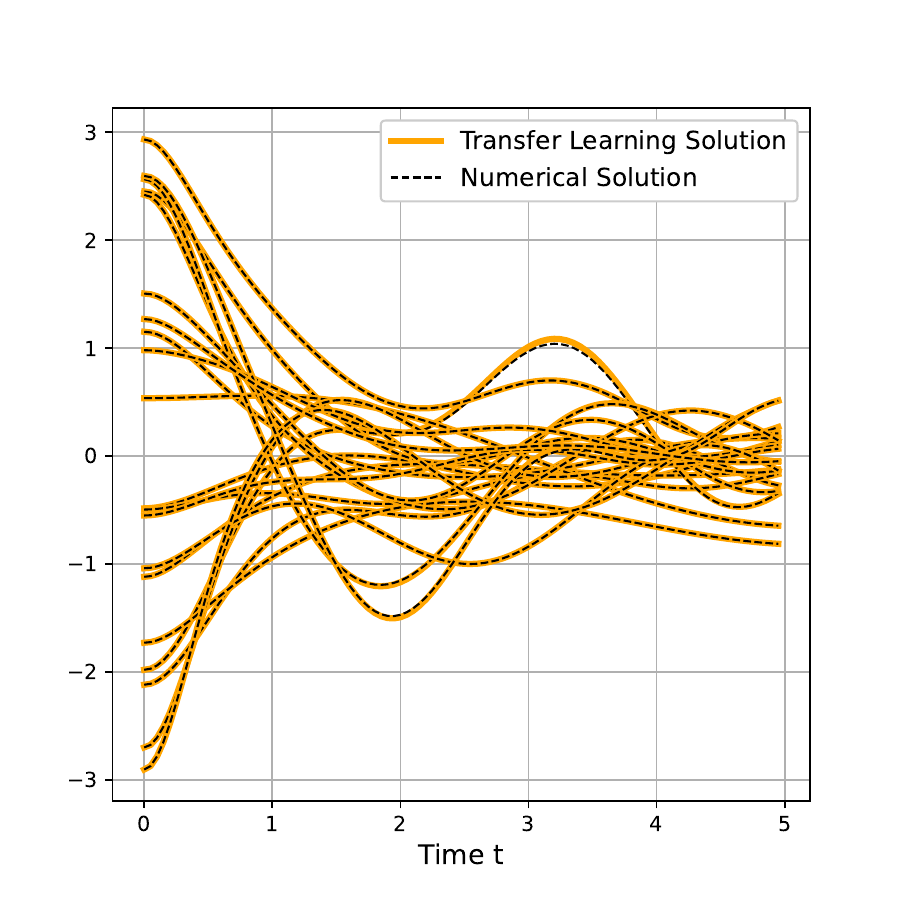}%
        \label{fig:b}%
        }%
    \caption{}
\end{figure}

We also test our method by comparing the transfer learning solutions with numerical solutions (explicit Runge-Kutta method of order 8) on 20 randomly generated Duffing equations in the same parameter range \ref{paraRange} (we fix $\beta = 0.5$ and $p=12$). Each Duffing equation can be solved in seconds. (Details in Appendix B) As shown in Figure 2(b), the 20 transfer learning solutions align almost perfectly with the numerical solutions, indicating that our methodology is very effective on equations in the same parameter range \ref{paraRange}.

\section{Conclusion}

 We introduced a framework using perturbation and one-shot transfer learning on PINNs to efficiently solve non-linear ODEs. We reduced non-linear ODEs to linear ODEs, trained a neural network with $k$ heads to handle them, and derived a formula for network weights. This approach allows us to solve various non-linear ODEs of the same form with a single trained network. Future work aims to extend this methodology to various non-linear ODE and PDE systems.

Our work should be considered as a starting point for this methodology. Future work is needed to extend the framework to non-linear ODE and PDE systems with various non-linearity forms.

\newpage
\bibliographystyle{abbrvnat} 
\bibliography{references}

\newpage
\begin{center}
    {\Large\bfseries Supplementary Material}
\end{center}
\begin{center}
    {\Large\bfseries Appendix A}
\end{center}

The matrix $B$ \ref{matrices} is an m by m diagonal matrix in which the first m-1 diagonals are all 1 and the last diagonal is $g_m$. Matrix $A$ \ref{matrices} is an m by m matrix whose second upper diagonal entries are all -1 and last row is $[g_0, g_1, ..., g_{m-1}]$.

\begin{equation}\label{matrices}
    B = \begin{bmatrix}
        I_{m-1 \times m-1} & \vec{0} \\
        \vec{0}^T & g_m
    \end{bmatrix}, 
    A = \begin{bmatrix}
        0 & -1 & 0 & \cdots & \cdots & 0 \\
        0 & 0 & -1 & \cdots & \cdots & 0 \\
        0 & 0 & 0 & -1 & \cdots & \vdots\\
        \vdots & \vdots & \vdots & 0 & \ddots & 0\\
        0& 0 & 0 & \cdots & 0 & -1 \\
        g_0 & g_1 & g_2 & \cdots & \cdots & g_{m-1}
    \end{bmatrix}.
\end{equation}

\begin{center}
    {\Large\bfseries Appendix B}
\end{center}

Here we show the diagram of the general multi-head PINN structure. In our real implementation to solve Duffing Equation, the network has 4 layers of hidden layers with width 256, 256, 256, 512 respectively. Hidden layers are all connected by $tanh$ activation functions. The activations of the last hidden layer is reshaped into a matrix $H \in \mathbb{R}^{2 \times 256}$. The matrix $H$ is connected to 10 heads of dimension 2 by a linear transform.

\begin{figure}[h]\label{diagram}
\caption{General Multihead PINN structure}
\centering
\includegraphics[width=0.9\textwidth]{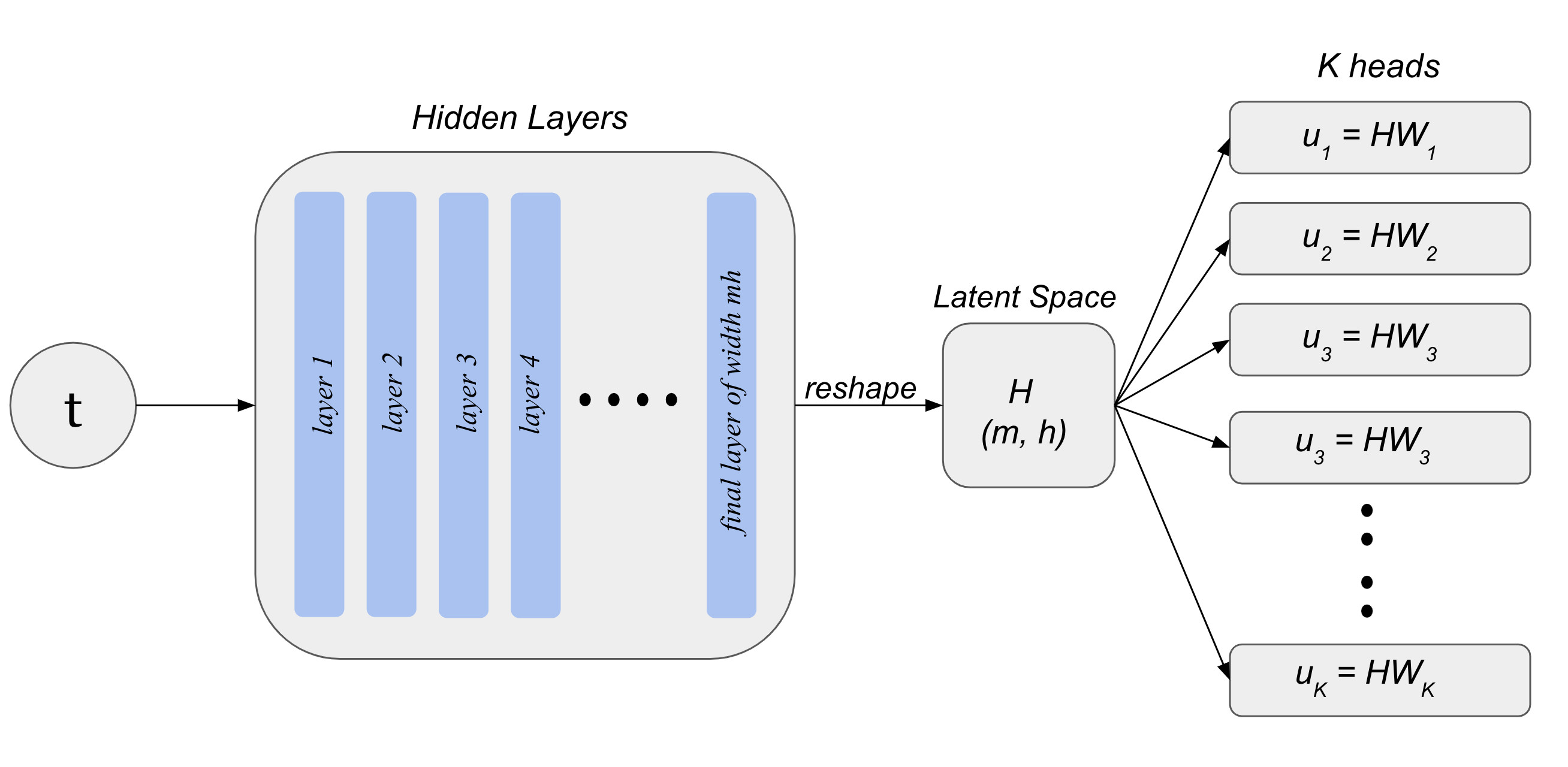}
\end{figure}

To train the network, we used Adam optimizer for 5000 iterations with an initial learning rate of $2\times 10^{-4}$. Note that we applied an exponential decay to the learning rate: the learning rate is multiplied by a factor of $0.96$ each 100 iterations. 200 random data points are uniformly generated in the domain $(0, 5)$ in each iteration as a sample set to compute the ODE loss. After 5000 iterations, the total loss (the sum of ODE loss and boundary loss) is reduced below $10^{-4}$.

We ran our code on google colab using an Intel Xeon CPU with 2 vCPUs (virtual CPUs) and 51GB of RAM. Solving an unseen Duffing equations generally takes $0.5p + 1$ seconds, where $p+1$ is the number of linear ODE systems we reduce to using perturbation method and the additional 1 s is the time to compute and invert the matrix $M$.

\begin{center}
    {\Large\bfseries Appendix C}
\end{center}

To reduce the ODE $Dx = f$ to first order, we introduce $m-1$ time-dependent variables: $\{ x^{(i)} \}_{i=1}^{m-1}$ to form a function $\bm{u}: \mathbb{R} \rightarrow \mathbb{R}^m$. $\bm{u} = [x, x^{(1)}, x^{(2)}, ..., x^{(m-1)}]^T$. $Dx = f$ is equivalent to:

\begin{equation}\label{expandedEq}
    g_0 x + \sum_{i=1}^m g_i \frac{d^i}{dt^i} x = f,
\end{equation}

The introduced variables are defined as: $x^{(1)} = \dot{x}$ and $x^{(i)} = \dot{x}^{(i-1)}$ for $i=2, 3, ..., m-1$. Expand these relations iteratively, we obtain $\frac{d^i}{dt^i} x = x^{(i)}$ for $i=1, 2, ..., m-1$, plug into \ref{expandedEq}, we obtain: 

\begin{equation}\label{lasteq}
    g_0x + \sum_{i=1}^{m-1} g_i x^{(i)} + g_m \dot{x}^{(m-1)} = f.
\end{equation}

The definitions of these $m-1$ variables introduces another $m-1$ constraints, together with \ref{lasteq}, we recover the linear ODE system in \ref{sysode}.

\end{document}